\documentclass[10pt]{article} 
\usepackage[preprint]{rlc}

\usepackage{amssymb}            
\usepackage{mathtools}          
\usepackage{mathrsfs}           
\mathtoolsset{showonlyrefs}     
\usepackage{graphicx}           
\usepackage{subcaption}         
\usepackage[space]{grffile}     
\usepackage{url}                
\usepackage{hyperref}

\title{Less is more - the dispatcher/ executor principle for multi-task Reinforcement Learning}

\RequirePackage{graphicx}

\author{%
  Martin Riedmiller\\
  Google DeepMind\\
  \texttt{\footnotesize riedmiller@google.com} \\
  \And
  Andrea Gesmundo\\
  Google DeepMind\\
  \texttt{\footnotesize agesmundo@google.com} \\ 
  \And
  Tim Hertweck\\
  Google DeepMind\\
  \texttt{\footnotesize thertweck@google.com} \\ 
  \And
  Roland Hafner\\
  Google DeepMind\\
  \texttt{\footnotesize rhafner@google.com} \\ 
}

\begin{document}
\maketitle


\begin{abstract}

Humans instinctively neglect irrelevant details when solving complex decision-making problems in environments with unforeseeable variations. This abstraction process appears to be a vital property for biological systems, helping to abstract away'' unnecessary details and boost generalization. In this work, we introduce the \textbf{Dispatcher/Executor (D/E)} principle for the design of multi-task Reinforcement Learning controllers. It proposes partitioning the controller into two entities: one that understands the task (the \textbf{Dispatcher}) and one that computes the controls for the specific device (the \textbf{Executor}) - connected by a strongly regularized communication channel. The core rationale behind this position paper is that changes in structure and design principles can improve generalization properties and drastically enhance data efficiency. It is, in a sense, a ``yes, and...'' response to the current trend of using large neural networks trained on vast amounts of data to rely on emergent generalization properties.
While we agree on the power of scaling - in the sense of Sutton's ``bitter lesson'' - we provide evidence that considering structure and design principles remains a valuable and critical component, particularly when data is not abundant and infinite, but a precious resource. A video showing the results can be found at \url{https://sites.google.com/view/dispatcher-executor}.


\end{abstract}


\section{Introduction}

Reinforcement Learning (RL) has evolved from single-task settings to multi-task scenarios, where a single agent must solve one of many potential tasks. Such requirements are ubiquitous in robotics, where a manipulator must handle diverse objects or a locomotion platform must navigate to varying goals. Multi-task challenges also arise in classical control, such as shaping fusion plasma \citep{degraveNature2022} or general set-point control \citep{hafnerriedmiller2011}. The prevailing approach typically relies on a monolithic neural network, conditioned on a task specification to modulate its behavior.

However, effective multi-task control demands two distinct types of knowledge: semantic task understanding (e.g., identifying which objects are relevant or what manipulation is required) and mechanical execution (e.g., specific knowledge of the device's kinematics and dynamics). Standard monolithic architectures conflate these requirements, forcing a single network to simultaneously learn general world knowledge and specific actuation skills.

In this work, we argue that multi-task RL benefits significantly from a bi-partite architecture that explicitly separates these concerns: a \textbf{Dispatcher} that parses the semantic intent, and an \textbf{Executor} that generates the control actions. Crucially, the communication interface between these modules should enforce a strict information bottleneck, filtering out task-irrelevant details. This forced abstraction allows the same Executor to be reused across a broad diversity of visual scenes, thereby boosting generalization—validating the principle that ``less is more.''

We formalize this Dispatcher/Executor (D/E) principle and present a concrete implementation within the domain of robotic manipulation. We provide extensive empirical evaluations demonstrating significant benefits across a range of tasks in both simulation and on a real robot. Notably, the D/E principle can also be applied retrospectively; we demonstrate this ``hindsight transfer'' by decomposing an existing real-world policy (trained for stacking specific colored blocks) into a D/E structure capable of general object stacking.

This position paper makes the following contributions:
\begin{itemize}
\item Introduces the Dispatcher/Executor (D/E) principle as a design paradigm for scalable multi-task RL controllers.
\item Proposes a concrete architecture for D/E in robotic manipulation, serving as both a proof of principle and a practical implementation guide.
\item Provides empirical evaluations in simulation and on real hardware, demonstrating substantial gains in data efficiency and generalization for multi-task RL.
\end{itemize}

\section{State of the Art}

The core claim of Rich Sutton's insightful 'Bitter Lesson' paper \citep{suttonBitter2019} is simple and compelling:
'general methods that leverage computation are ultimately the most effective'.
This has been demonstrated several times in the history of AI, e.g. in game playing \citep{silver2016mastering}, 
computer vision \citep{KrizhevskySH17}, or more recently, in large language models \citep{VaswaniSPUJGKP17}: huge amounts of data trained in huge neural networks
with huge amounts of compute lead to unprecedented and surprising generalisation \citep{openai2023gpt4}, while the underlying training principle is general and relatively simple.

Given the big success in language, it is evident to investigate, 
if the same principles of scaling up to large amounts of data and
the use of large networks show the same generalisation effects in control. The GATO~\citep{reed2022generalist} approach demonstrated that
a reasonably large transformer style neural network is able to show good control performance with a single network 
on a variety of heterogeneous tasks from Atari to robotics. RoboCat~\citep{bousmalis2023robocat} used a related approach focusing on robotic control with the final goal to train a versatile model that can transfer to arbitrary new tasks. RT-1~\citep{brohan2022rt} and RT-2~\citep{zitkovich2023rt} start from a pre-trained language model and therefore have a built-in semantic understanding of tasks and the environment.
All of the above models are trained on vast amounts of data using a transformer-style controller architecture with an
additional input pathway to communicate the task. Since these methods are by default extremely data-hungry, data-augmentation is one path to generate even more data out of the data collected \citep{yu2023scaling} and by this enforce generalisation within the distribution of the data generated.

In \cite{ahn2022i} it is shown that a pre-trained language model and a set of pre-trained skills can be efficiently combined by learning to sequence these skills, based on the assumption that language model and the skills share some common understanding of the main task. While this gives some form of general task execution in known environments, a large amount of data has to be used to make sure the pre-trained skills will be able to execute on a sufficient enough set of variations of the environment.

Another scheme that is very common for data efficient multi task approaches is to learn some form of task embedding that allows to generalise in the task space (e.g. \cite{hausman2018learning, nachum2018dataefficient}).
While some of these ideas can be combined with the D/E approach, these methods solely focus on the command part in the communication channel. Therefore the expected generalisation by abstracting the observation and the command, as we demonstrate here, is missing.

While we agree that scaling is one way to achieve more and more general controllers, we at the same time believe that even large scale systems should be 'data-efficient', i.e. should make the best use of the data available. The core thesis of this paper is that incorporating structure into the architecture and the communication channel can drastically enhance the
generalisation capabilities 'by design'. In particular, enforcing the right representation within the controller seems to be an important enabler. Recent examples of this line of research can be found in \cite{levine2016endtoend, pinneri2023equivariant, groth2021goal}.

The idea of a two (or more) level architectures for control has been explored in many different variations (e.g. \cite{vezzaniSkills2022, zhang2021hierarchical, riedmiller2018learning, hafner2022deep}. The D/E principle intentionally defines the separation from the viewpoint of the type of knowledge required to run the components: a dispatcher component that understands the task through general world knowledge (that can be trained on general data like text, images or videos), 
and an executor component that specifically understands the device (trained through active interaction with the device). 
The restrictions on the communication channel enforce an abstract information exchange between these types of knowledge that boosts generalisation. We believe that in particular in this new era of large language models that are able to represent general world knowledge, this structural principle can play an important role to boost learning control on real robots, where data-efficiency with respect to device interaction is a key requirement.

\section{The Dispatcher/ Executor Principle for Multi-task RL}

\subsection{The general idea}

\begin{figure}[htbp]
  \centering
  \includegraphics[width=420pt]{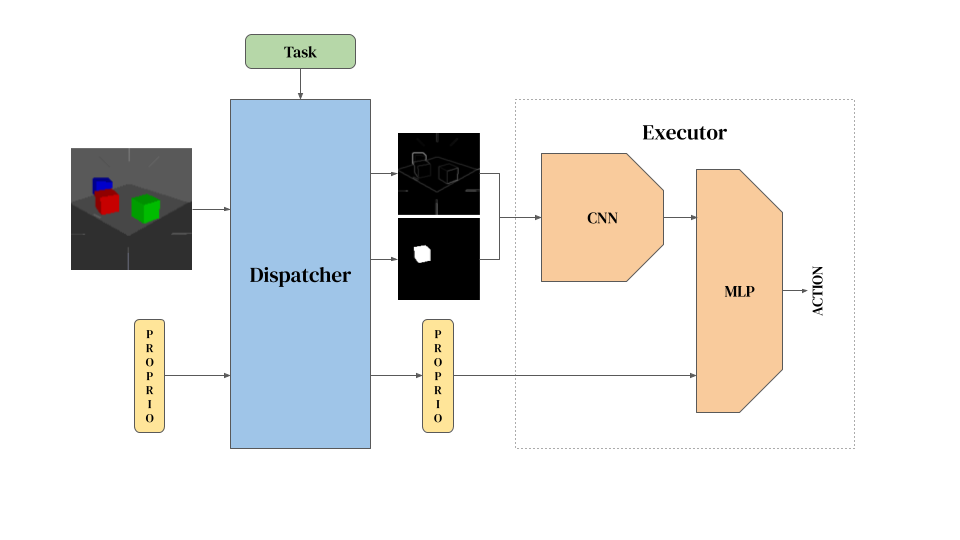}
  \caption{Illustration of the D/E architecture. The controller consists of two modules, the dispatcher and the executor. The dispatcher gets the current observation and modifies it according to the task description. This abstract message is sent to the executor, which computes the according action for the device.}
  \label{fig:DE}
\end{figure}

The proposed dispatcher/ executor (D/E) principle for control makes two main points: 1. separate the control architecture into two entities according to the type of knowledge they require (dispatcher and executor), and 2. define a suitable communication channel between those two modules with strongly regularizing properties.

The two parts are defined through their specific duties: the dispatcher is responsible for semantically understanding the task and calling the executor with the respective instructions. 
The executor is responsible for computing the actual control signal that drives the device to achieve its goal communicated from the dispatcher. Thus, whereas the dispatcher needs an understanding of the world in general to be able to understand the task, the executor needs an understanding of how the device reacts to its control signals. Whereas in this paper we mainly focus on a single dispatcher and a single executor, in its generalized form, the controller might have several executors that are specialized on realizing different skills. E.g., to fulfill a complex task, the dispatcher might call several executors in a row.

The second crucial point of the D/E principle is the design of the communication channel between dispatcher and executor. It has two relevant properties: firstly, it should enforce compositionality of the transmitted command information, and secondly, it should reduce the transferred information to the minimum that the executor needs to successfully perform the task. The requirement of compositionality might be achieved by forcing some kind of structure into the communication - e.g. in form of a certain artificial language - that forces the exchanged information to demonstrate some kind of abstraction.

Enforcing this language-like communication we imagine as a property that is incorporated into an agent through structural design (e.g. found through evolution in a natural agent), and finding the right words of this language is something that finally is learned by the agent through experience and shaped through an according loss function.

It is beyond the scope of this work to figure out, if and how such an interface can be found through evolution and/or learning. We will instead focus on proposing a concrete implementation of a D/E architecture and the 
communication interface in the spirit of an existence proof and show the concrete benefits on the example of robotic manipulation.

\subsection{A concrete D/E implementation for robotic manipulation \label{sec:DEmani}}

Typical tasks in robotics involve the manipulation of 
the environment, like e.g. 'push the button', 'lift the screw', 'put the bottle on the table', ...
Following the D/E principle, we suggest the following concrete implementation (see figure \ref{fig:DE}):
the dispatcher receives the task description and the raw sensor observations at each time step, which 
we assume includes camera images and optionally other sensor information, e.g. proprioceptive values.
The dispatcher analyses the task description, infers what to do and identifies the targeted objects in the
image. It then encodes (reduced) information 
about the target objects and environment as arguments for the executor.
The message to the encoder takes the following form: 
$ExId, arg_1, \ldots , arg_n$ where $ExID$ identifies the executor and $arg_1, \ldots, arg_n$ contain separated and compressed information about target object(s) and scene.

There are different ways now to specify a targeted object: an obvious way would be through 
6D pose coordinates, which of course are only available if we have ground truth, which is most often not
the case. Since the current scene is represented through one or more camera images, we instead suggest to 
mark object information in pixel space. Concretely, we suggest to 
code the information about the target object through a simple masking operation: all pixels in the 
observed image that belong to the object are classified with 1, the rest is set to 0.
This operation reduces the information drastically 
to what the executor might need to locate the object and infer pose and shape if needed. Note, that this
information now abstracts away further properties of the object, like type or color. The executor 
therefore works immediately for a broad range of objects that may appear very differently in the original observation space (here an RGB image). We call this the 'mask' filter of a target object for further reference.

In addition, some general information about the scene might be required as well. Think of a target object lying close to another object - a situation which may require special handling. In the sense of reducing information to a minimum required to solve the task, we here
suggest to run the full image through an edge detector and provide 
the result as a further argument to the executor (referred to as 'edge' filter in the sequel).

An alternative representation of target objects can be done through 'pointing' to the object: A simple implementation of this is to compute the pixel centroid of the mask of the object and then represent this through a 5x5 blob at the respective location in the image. Again, all other pixels are set to 0. This coding scheme is called the 'pointer' filter in the following.

The above operations of identifying target objects and providing background information are applied to all camera images available, to have full information of the scene. 

We admit that the proposed encoding scheme is very much 
tailored to the typical tasks that occur in robotic manipulation, and the proposed filter operations are just a small selection from many possible choices. This is intended, since here we mainly want to give a concrete 
example of the existence of such an encoding and its usefulness in a certain application. We are convinced that
the idea behind the approach transfers to other application scenarios as well, even more, that future work
will aim at finding these coding schemes through learning rather than through design.

\subsection{Workflow and learning within the D/E architecture for manipulation}

The overall workflow in the proposed D/E architecture is as follows: the dispatcher gets the full observation vector plus the task description as the input. It parses the task description and identifies the objects involved. Then it computes the respective pixel frames for the arguments (e.g. using one or more of the above filter schemes, masks, pointers, edges). This information is then transferred to the executor in every control step.

Learning in a D/E architecture is currently restricted to the executor. Learning is not different from the standard monolithical controller, apart from the fact, that the input is represented differently. The executor can therefore be trained in many different ways. In the following we investigate Reinforcement Learning from scratch as well as the distillation of a previously learned policy to a D/E architecture in a student/teacher setup. Another interesting direction, which is beyond the scope of this paper, is the investigation of learning multiple variants of representations in an off-line RL setting of pre-collected data \citep{lange2012batch, riedmiller2021collect}.

\section{Empirical investigations of targeted properties}

\begin{figure}[htbp]
\begin{minipage}{.24\textwidth}
  \centering
  \includegraphics[width=1.0\linewidth]{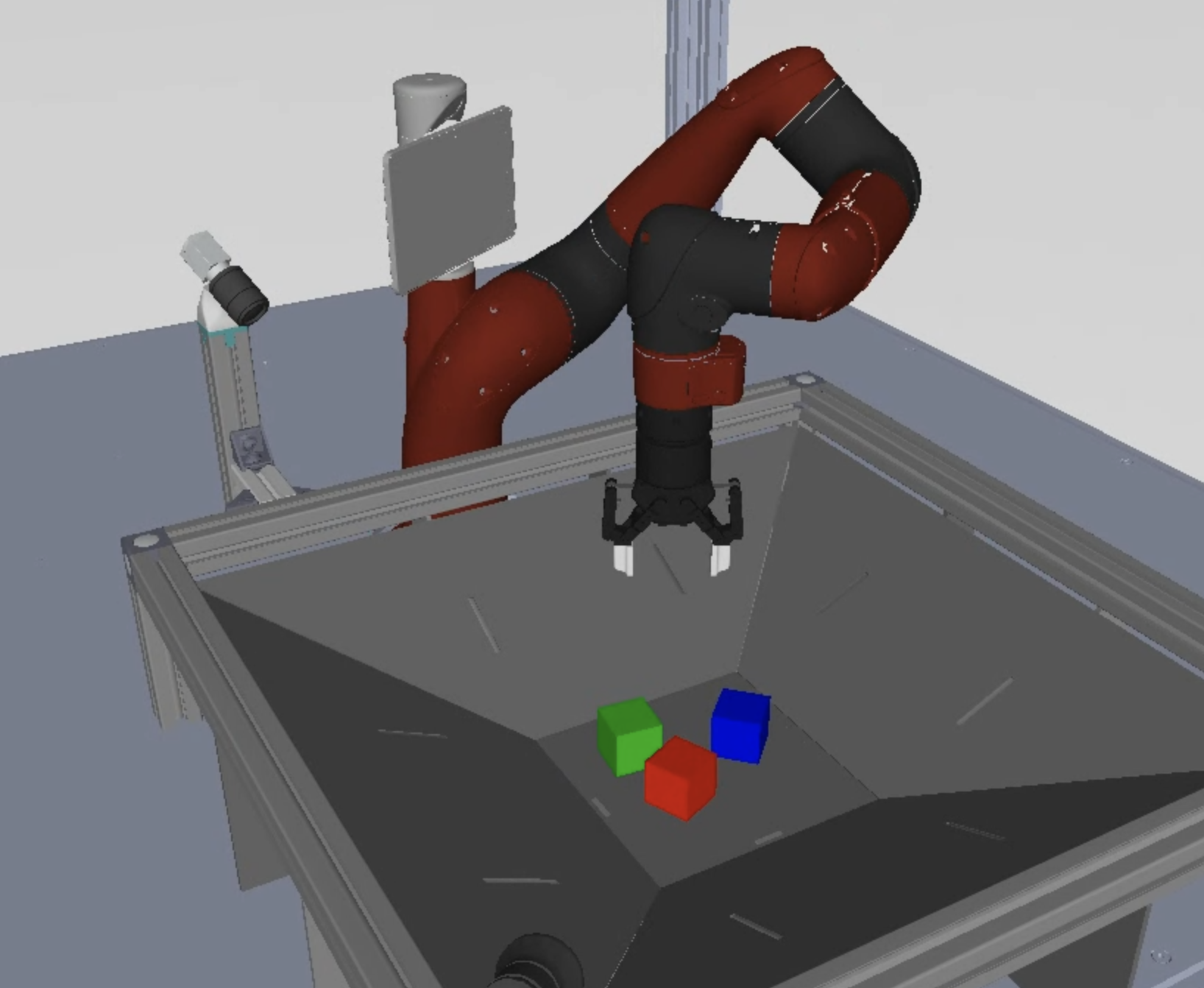}
\end{minipage}
\centering
\begin{minipage}{.24\textwidth}
  \centering
  \includegraphics[width=.8\linewidth]{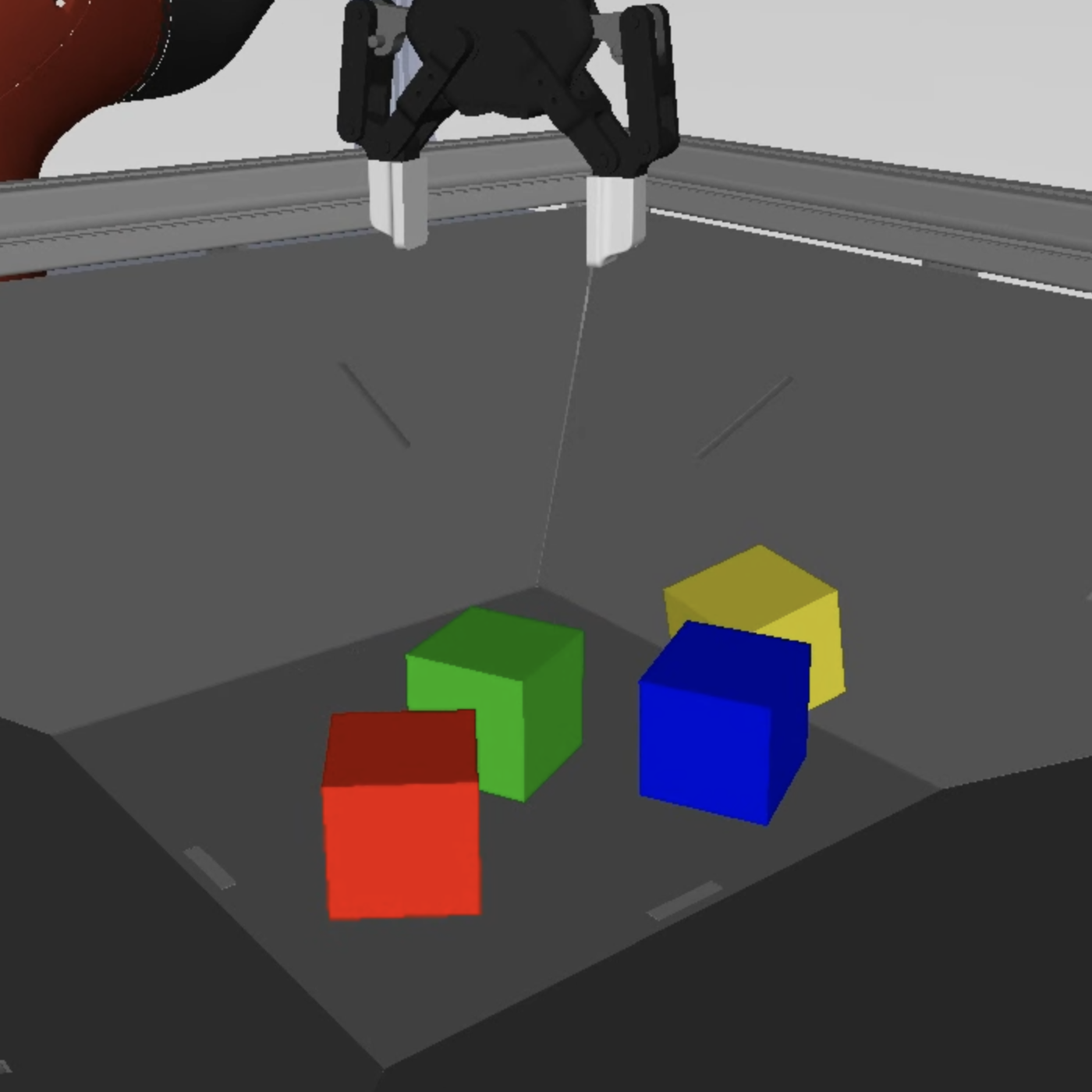}
\end{minipage}%
\begin{minipage}{.24\textwidth}
  \centering
  \includegraphics[width=.8\linewidth]{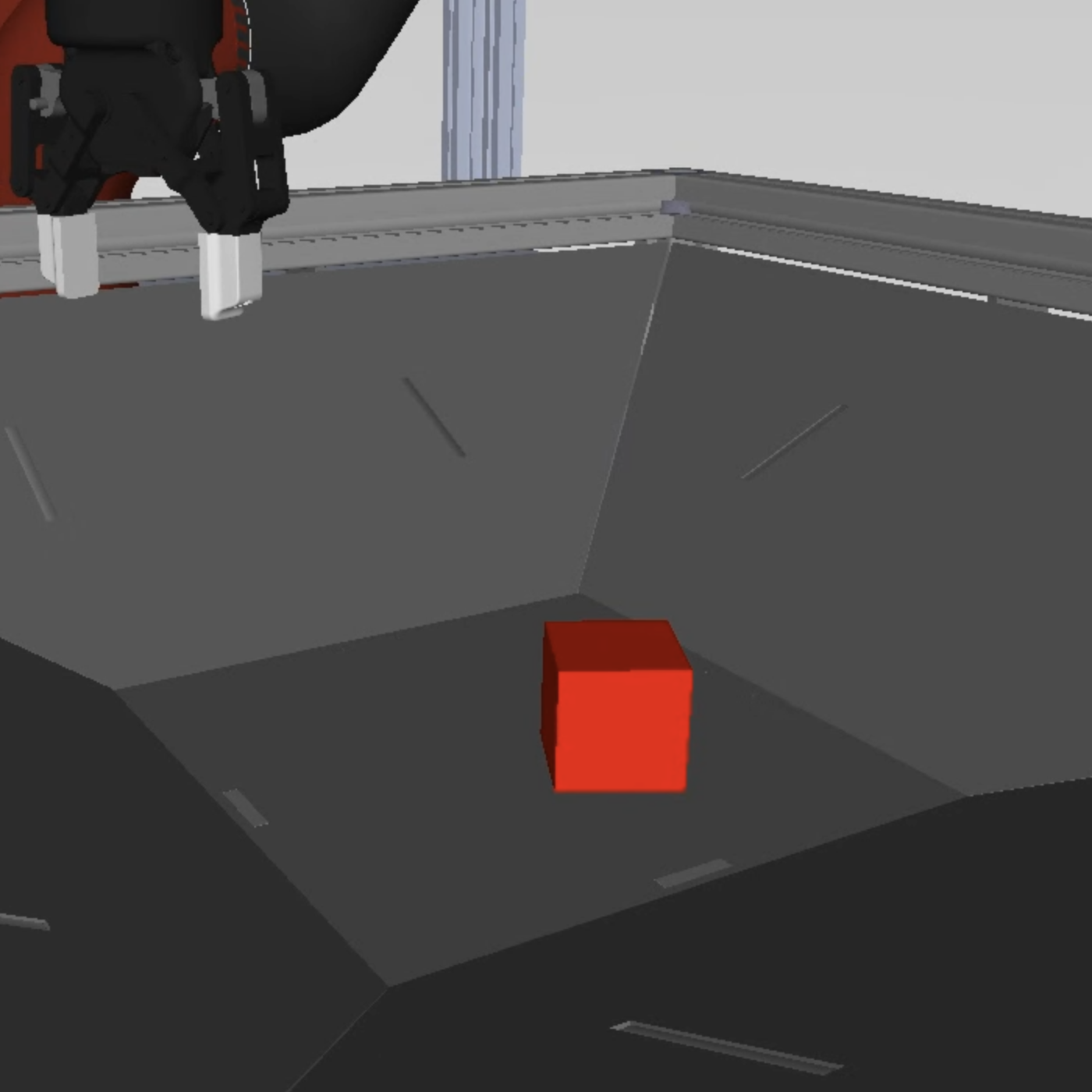}
\end{minipage}
\begin{minipage}{.24\textwidth}
  \centering
  \includegraphics[width=.8\linewidth]{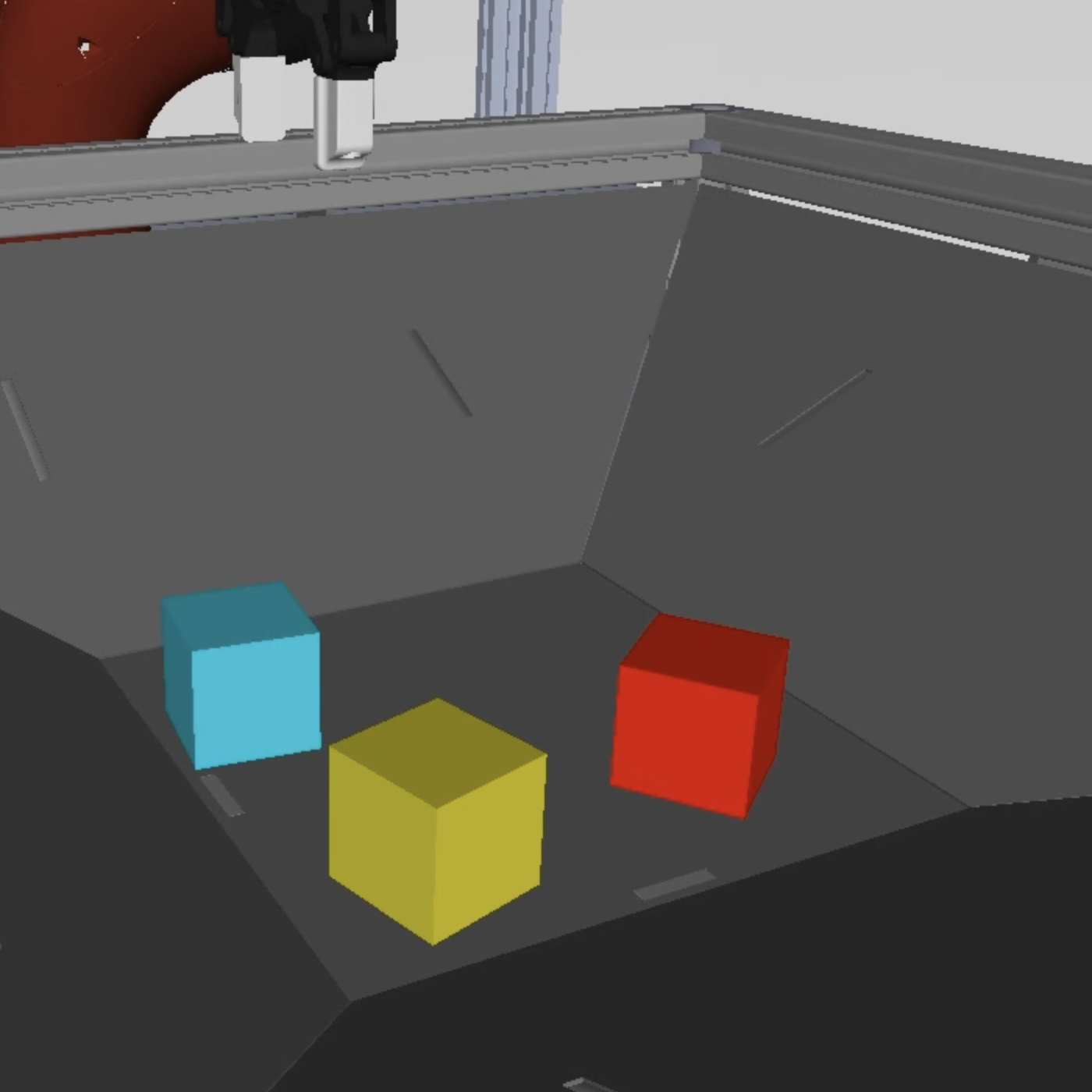}
\end{minipage}
\captionof{figure}{Simulation setup. From left to right: original setup with 3 objects, "Four cubes", "One cube", "Recolor". The controllers are trained to lift the red object (single task), the red, left or green object (multi-task), and evaluated also in non-training situations.}
\label{fig:simenv}
\end{figure}

In this section, we highlight some properties of the D/E architecture on various learning
experiments. It is not meant as an in-depth empirical evaluation, it rather gives exemplary 
insights, of how the proposed D/E structure of the controller can improve learning and generalisation
properties in comparison to a classical monolithic control structure. We provide experiments both
in simulation and on the real robot on two set of tasks, lifting an object (in simulation)
and stacking of two objects (real robot).
If not otherwise noted, 'standard' refers to a classical, monolithical neural network structure which is extended
by an extra-input for task communication.

The dispatcher-executor architecture (see figure \ref{fig:DE}) uses a hardcoded dispatcher and a learned executor.
The dispatcher gets the task specification (e.g. 'lift green object' or 'stack red on blue object') 
and then segments the according target objects - in this case using simple color segmentation (however, it is straightforward to replace this by more capable segmentation schemes, e.g. \cite{oquab2023dinov2}. In addition, it filters
the whole scene using an edge operator. Then, the executor is called, providing the respectively filtered images
as separate arguments. If not otherwise noted, 'D/E' refers to the default setting of using a segmentation mask
for the objects, and the edge operator for the overall scene (as described in \ref{sec:DEmani}).

For training both the standard architecture and the executor in the D/E case, a distributional variant of MPO \cite{abdolmaleki2018maximum}, with a Mixture-of-Gaussian critic was used. The agent receives the images of three cameras placed around the robot's workspace, as well as proprioceptive observations. Both architectures employ a three-block ResNet that computes an eight-dimensional embedding vector for each pixel input. These embeddings are concatenated with all proprioceptive observations and fed through a three layer MLP for computing the agent's five-dimensional actions.

\subsection{Single task: lifting the red cube (simulation) \label{sec:single}}

\begin{figure}[htbp]
  \centering
  \includegraphics[width=300pt]{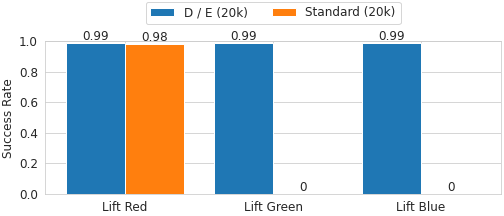}
  \caption{Single task training: 'lift red'. Both standard architecture and D/E architecture are trained for 20,000 episodes. Figure shows evaluation results for different tasks. The standard architecture solves the learned task only, whereas the D/E architecture can also lift green and blue cubes}
  \label{fig:liftred}
\end{figure}

For the first experiment, the scene consists of three cubes with different colors (red, green, blue; see figure \ref{fig:simenv}).
The single task setup is to learn to lift the red cube (blue and green cubes are mere distractors).
Figure \ref{fig:liftred} shows the results. Firstly, both approaches
(standard and D/E) need about 20,000 episodes to train the task successfully (98 \% success in evaluation),  i.e. the number of training episodes to get to a successful 
'lift red' policy does not change significantly with the controller structure used.

Transfer to new tasks: since the input representation of the executor abstracts away the color of the cubes, the D/E architecture can transfer immediately to the tasks lift green or lift blue by setting the executor input respectively through
the dispatcher module. Therefore the D/E controller is able to additionally perform 'lift green' and 'lift blue' with
similar success rates - without any further training and data collection. The standard monolithical architecture is specialized to the single task and lacks the capability of transfer.

Findings:

\begin{itemize}
    \item for a single task, number of training episodes for standard and D/E architecture are in the same range
    \item the D/E architecture can immediately transfer to new tasks ('zero effort transfer')
\end{itemize}

\subsection{Multi-task 1: Lifting one out of three cubes (simulation) }

Like above, the scene consists of three cubes, and now the training task is to alternately lift one of the cubes specified by its color.
In the standard
controller, we specify which cube to lift through an additional input.
For the D/E controller, the dispatcher was adjusted accordingly and is now selecting the respective 
representation of the targeted object and providing the respective information to the executor.

\begin{figure}[htbp]
  \centering
  \includegraphics[width=370pt]{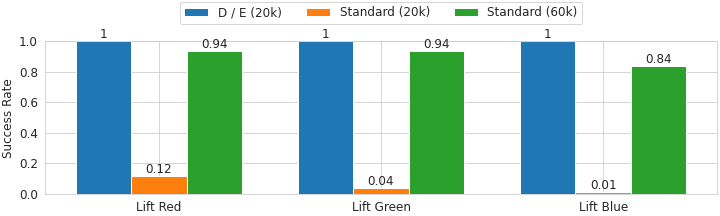}
  \caption{Multi task training: 'lift red/ green/ blue'. After 20k episodes, the D/E controller masters all three tasks, whereas the standard controller still performs purely (orange). After 60k episodes using 3 times as much training data, the performance of the standard controller (green) increased significantly, but still does not reach the D/E results.}
   \label{fig:liftmulti}
\end{figure}

The results are shown in figure \ref{fig:liftmulti}. 
The D/E architecture can solve the three tasks in about 20k episodes, as in the single task setup before. This is due to the fact, that for the policy, the control task remains the same, no matter, which cube needs to be lifted. It therefore can fully leverage the data of all tasks and learn the lifting task quickly. On the other side, the standard architecture after 20k episodes, still sees pretty low success rates (between 1 and 12 \%). It needs much more experience to solve the three tasks, and only after 60k episodes, reasonable success rates are achieved. We even had to increase the network size in comparison to the single task experiment to get to this result.

Findings:

\begin{itemize}
    \item in the multi-task setup, the D/E architecture can leverage the abstraction and learn much more efficiently than
    the standard architecture. It can focus on learning the control task rather than the distinction between the tasks and
    this gives a dramatic speedup in learning.
\end{itemize}

\subsection{Multi-task 2: Lifting one out of three objects with multiple shapes (simulation) }

\begin{figure}[htbp]
  \centering
  \hspace{15pt}
  \includegraphics[width=212pt]{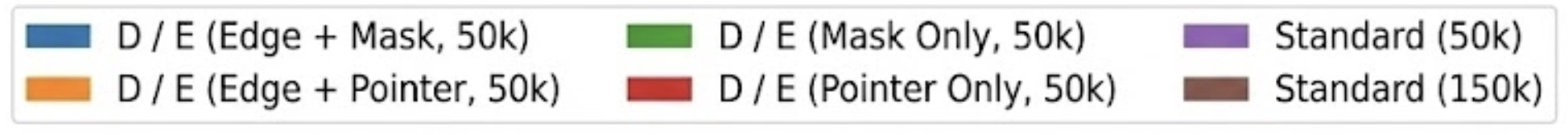}
  \includegraphics[width=400pt]{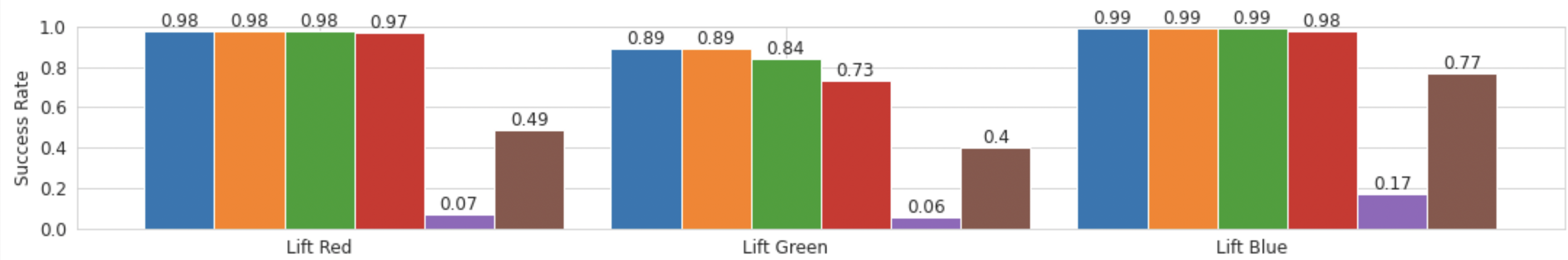}
  \caption{Multi task training: 'lift red/ green/ blue' with varying object shapes are used. D/E architecture with different representations performs well after 50k episodes. Standard architecture has low performance after 50k episodes and even after 150k episodes is worse than D/E.}
  \label{fig:liftgeneral} 
\end{figure}

\begin{figure}
\centering
\begin{minipage}{.49\textwidth}
  \centering
  \includegraphics[width=.8\linewidth]{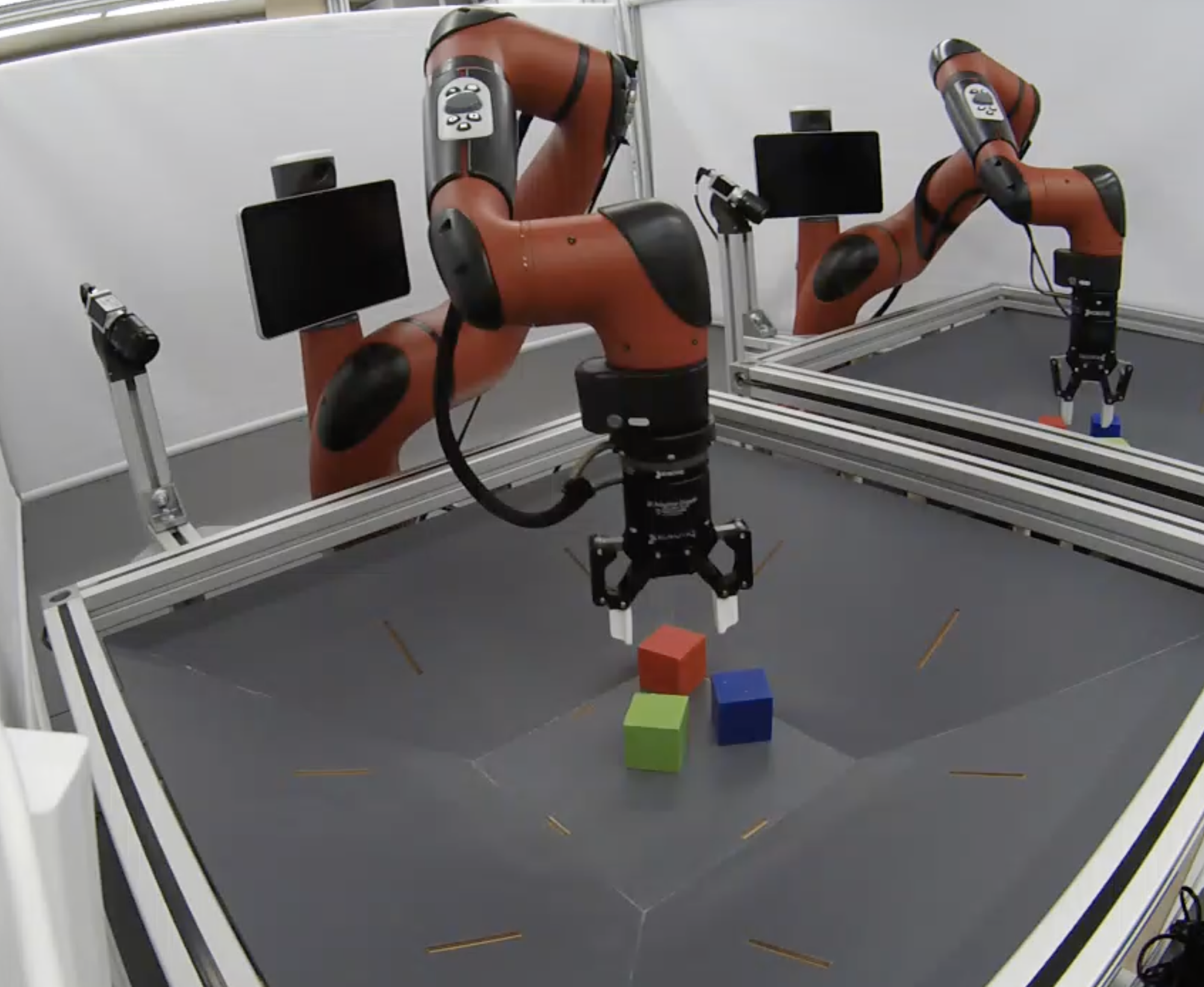}
\end{minipage}
 \begin{minipage}{.49\textwidth}
 {
   \centering
   \includegraphics[width=.7\linewidth]{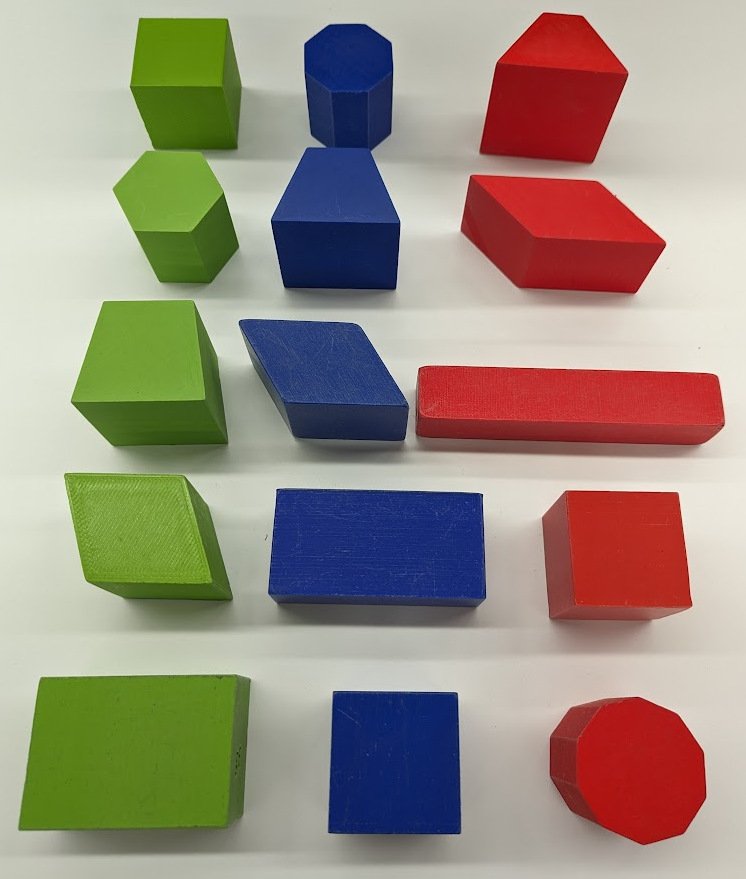}
    }%
  \end{minipage}%
  \captionof{figure}{Left: the real robot environment. Right: the RGB sets 1-5 (top to bottom). Original task is to stack the red object on the blue object, extended task is to stack any object on any arbitray object of one set, which evaluates generalisation over color and shape.}
  \label{fig:setup_real}
\end{figure}

One critical question with a reduced representation is, whether the carried information is still rich enough 
to cope for desired variances in the task setup. To look a bit deeper into this, in the following the
objects are additionally varied in shape. This problem is considerably more complicated than grasping cubes, 
forcing the controller to deliberately choose grasping points depending on the object shape. For example,
a rectangular shaped object needs grasping points different from a triangular shaped one. For this experiment, we used the objects of the RGB stacking benchmark \citep{lee2021beyond}, see figure \ref{fig:setup_real}, right.
Please note, that the distribution of shapes is different for different colors. This explains the varying success rates for different colors, e.g. the green objects are more challenging to lift in average.

Results are shown in figure \ref{fig:liftgeneral}. We tested four different variations of representation for the D/E architecture: the typical setup with mask and edge filter, then a setup with pointer and edge filter, and both a mask filter and pointer without scene information. All setups are trained for 50,000 episodes. Both mask and edge as well as pointer and edge work very reliably and lead to comparable results. This indicates, that there is some freedom for choosing a good representation. One can see some performance drop, if only the target object is communicated - this drop is more significant for the pointer only case (which has fewer information) than for the mask only setup. The standard architecture performs significantly worse - after 50,000 episodes only between 6 \% and 17 \% success rate is achieved. Using three times more data, after 150,000 episodes the network performs better, but is sill significantly worse than the D/E architecture. Again, this stresses the hypothesis, that bringing in structure can drastically improve the learning process.

Findings:

\begin{itemize}
    \item the reduced representation of the D/E architecture should be kept rich enough, such that it enables the policy to adapt to varying requirements. 
    \item different representation schemes might lead to successful behaviour
\end{itemize}

\subsection{Robustness with respect to scene variations}

\begin{figure}[htbp]
  \centering
  \includegraphics[width=250pt]{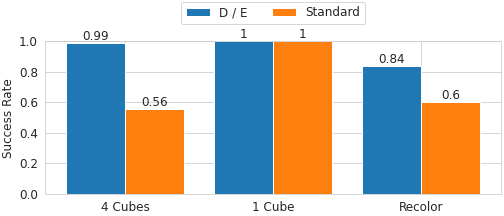}
  \caption{Evaluation of robustness. The controller is trained and evaluated on 'lift red'. Evaluation happens on varying background scenes, with only one cube in the scene, with four cubes or with re-colored distractor objects.}
  \label{fig:liftscenes}
\end{figure}

One potential challenge in rich scene representations is that the controller might concentrate on task irrelevant
features and rely on them for successful execution. We expect, that a reduced representation as it is used
in the D/E architecture, will be more robust to variations in the setup of the task.
As a straightforward experiment we investigate, how the single task controllers trained in section \ref{sec:single}
react to slight changes in the setup. In particular, we investigate the performance, when changing the number
of objects in the scene from three to one or four respectively, or change the color of the distractor objects.

As shown in figure \ref{fig:liftscenes}, both the standard controller and the D/E architecture are robustly performing
in the '1 cube' case. If instead four objects are put into the scene, the standard controller significantly looses performance,
whereas the D/E architecture still reliably solves the task. Similar behaviour can be observed when the distractor objects have different colors.

\newpage
Findings:
\begin{itemize}
    \item the reduced representation in the D/E architecture shows more robust behaviour when the environment conditions change.
\end{itemize}

\subsection{Real Robot Experiments - Teacher/ Student setup}

\begin{figure}[htbp]
  \centering
  \includegraphics[width=420pt]{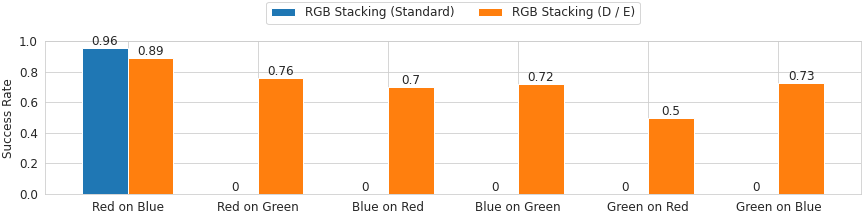}
  \caption{Real robot experiments: Cloning a 'red on blue' policy to the D/ E architecture. The teacher performance was an 96 \% success rate on average for 'stack red on blue' of the RGB stacking task}
    \label{fig:stackreal}
\end{figure}

To add another angle of use cases of the D/E architecture, we examine the use of a D/E controller
in a student-teacher setup. Concretely, we take a policy that was trained to solve a single task on a real robot ('stack red object on blue object') and distill it to a controller following the D/E principle.
A successful transfer will allow to open up a whole class of tasks ('stack object X on object Y'), without collecting
any further training data - a potential huge gain in data-efficiency.

The dispatcher in this case is restricted to find colored objects and advise the executor accordingly
(by setting the right mask), but it is obvious how to extend this to more general objects, e.g.
exploiting the recent advancement in general segmentation algorithms, e.g. \citep{oquab2023dinov2}.

As a teacher policy, we took a policy that has been fully self-learned on real robot experience to stack the
red object on the blue object on the five different object sets of the RGB stacking benchmark \citep{lampe2023mastering}. Note that this was learned from ground off directly from pure experience and no additional simulation data or tele-operation data was used.
The performance of the original policy was about  96\% in average over all 'red on blue' object sets 1-5. 

For evaluation of the distilled policy, we did two sets of experiments, one based on similar cubes and one on the RGB sets 1-5, where each object has a different shape. The results are shown in figure \ref{fig:stackreal}. For cubes, the stacking performance achieved high success rates over all color combinations. For the RGB sets, 'red on blue' achieves the highest result close to $90 \%$. Satisfyingly, not much performance was lost in the distillation process. Also for the other color combinations, average success rates of $50 \%$ or over are achieved.
This is a strong result, given that the shapes in the different sets are quite challenging and stacking these combinations has never been seen during the training phase of the executor. The distilled D/E policy gives a powerful baseline for finetuning the policy to perfection in all combinations.

Findings:

\begin{itemize}
    \item the D/E principle can be applied in student/ teacher scenarios, boosting existing policies to novel and more general application scenarios without the need to collect more data on the task.
    \item this highly capable stacking controller has been entirely self-learned from experience collected on the real robot.
\end{itemize}

\subsection{The power of the dispatcher}

\begin{figure}[htbp]
\centering
\begin{minipage}{.49\textwidth}
  \centering
  \includegraphics[width=.6\linewidth]{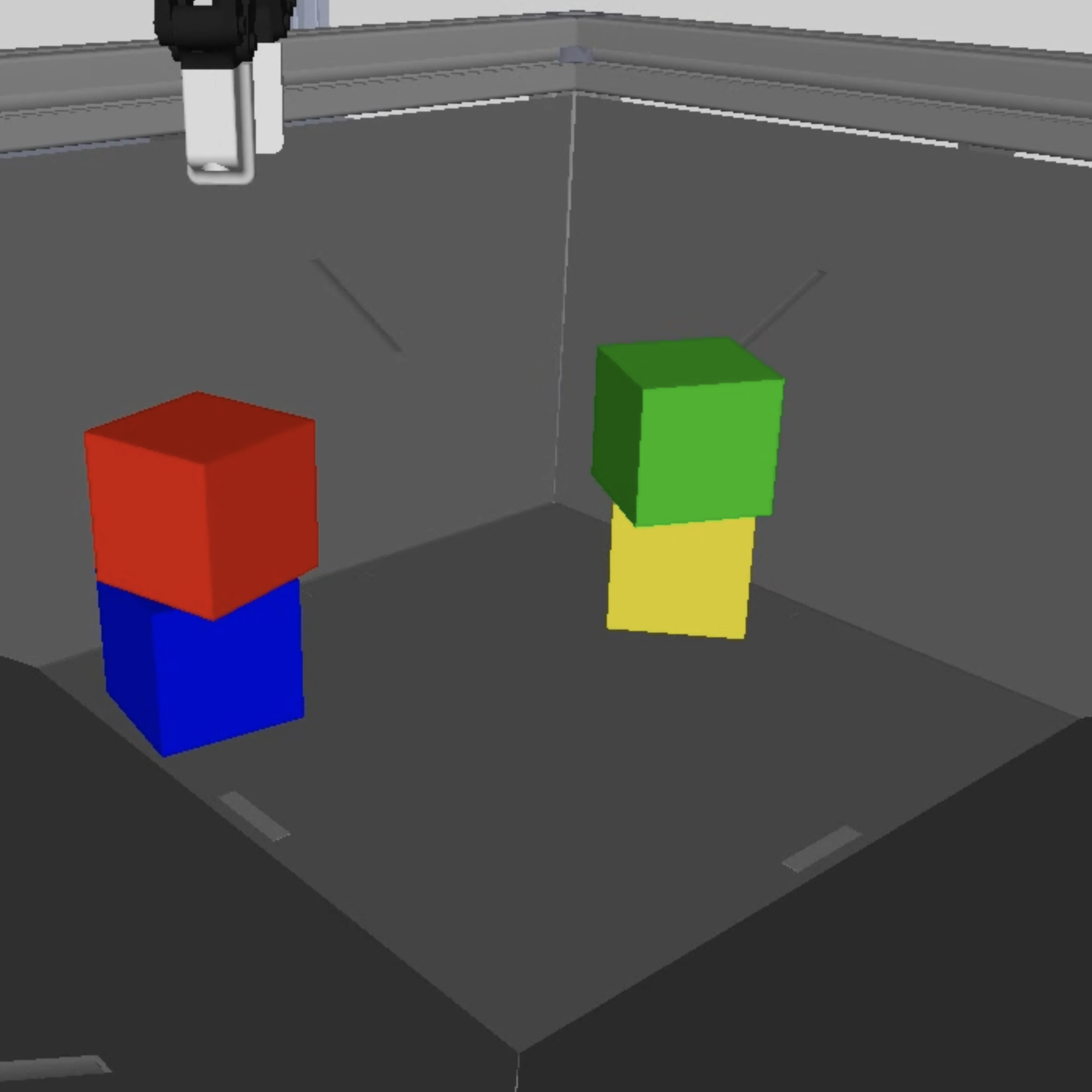}
\end{minipage}%
\begin{minipage}{.49\textwidth}
  \centering
  \includegraphics[width=.6\linewidth]{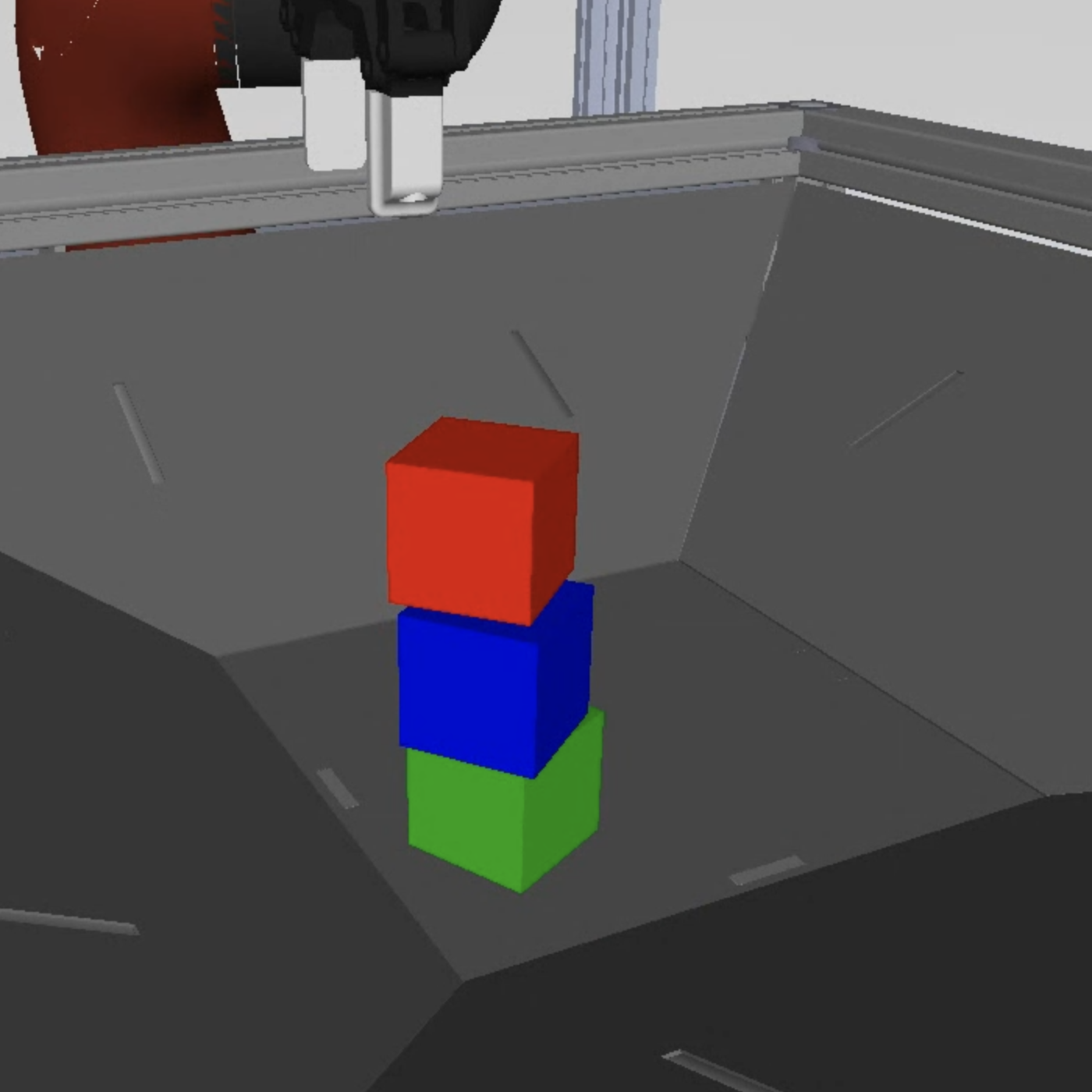}
\end{minipage}
\captionof{figure}{Zero-shot transfer of a D/E stacking policy to unseen task "Two Towers" (left) and "Triple Stack" (right) through adaptation of the dispatcher.}
\label{fig:dispatcher_eval}
\end{figure}

The dispatcher is used to understand the task description and translate it into the call of the executor with respective arguments. This is a versatile mechanism - for example, the dispatcher can call multiple executors sequentially in order to fulfill more complex, and long horizon tasks.

To demonstrate this, we are looking into two tasks: two-towers and triple-stack.

For two-towers, we ask to build two separate towers out of 4 cubes in the workspace. The executor is a simple stack X on Y policy, trained according to the D/E principle with masks and edges. 
The dispatcher identifies the four objects and then sequentially calls stack object 1 on object 2 followed by stack object 3 on object 4. Here, the different executors are just called after a time limit is reached, but more complicated schemes are possible and will be examined in future work. Note that no further executor training is needed to achieve this. 
The immediate success rate is $46 \%$. The main failure case is when the executor destroys the first tower when building the second, which is something it has simply not been trained to avoid. Finetuning the executor to respect existing towers will increase the success rate.

Triple stacking means stacking three objects on top of each other. This has been tried by other efforts using a single policy \citep{bousmalis2023robocat}. Here, we modify the dispatcher and ask to stack object 1 on object 2, followed by stack object 3 on object 1. Again, the executor is the basic D/E stacking policy. The observed success rate is $38 \%$. The main reason for failure here is that the second stack is not high enough (this was not seen during the original training phase). However, the initial performance is strong, and finetuning the executor with accordingly collected data will solve this task.

\subsection{Open-Vocabulary Semantic Dispatcher}
\begin{figure}[htbp]
\centering
\includegraphics[width=1.0\textwidth]{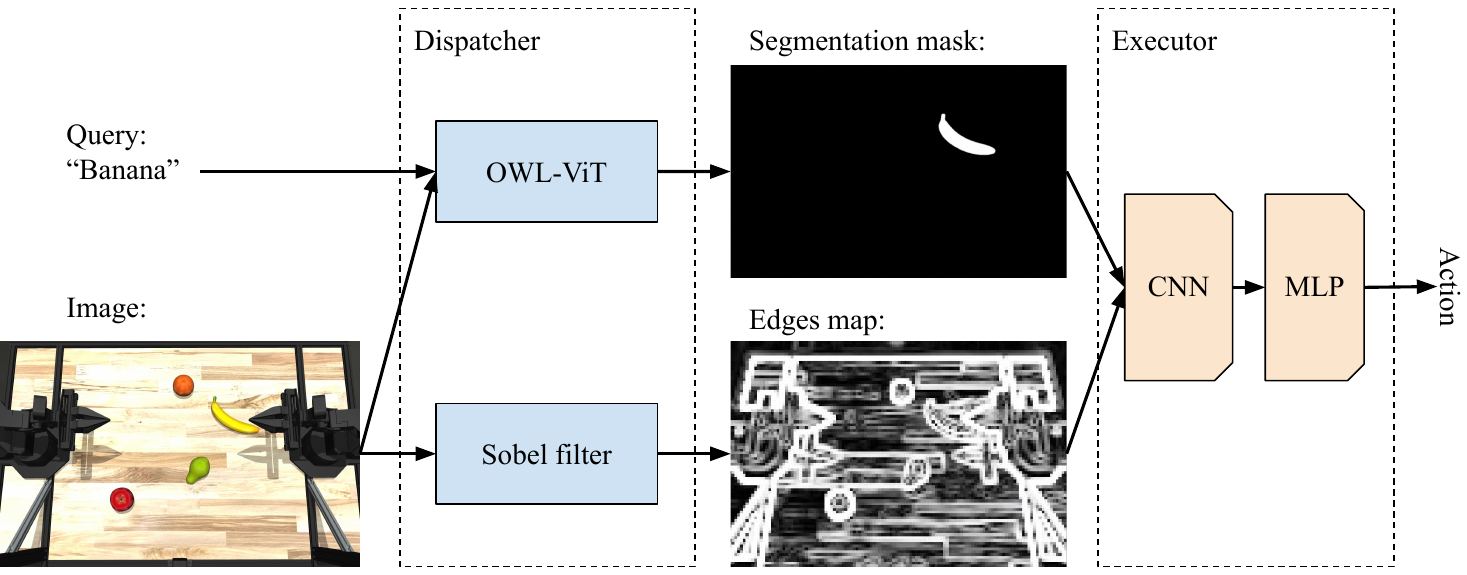}
  \caption{Overview of the open-vocabulary Dispatcher-Executor architecture. The generic Dispatcher processes camera inputs through two parallel streams: a semantic stream using the pre-trained OWL-ViT to generate high-level segmentation masks based on natural language queries, and a structural stream using standard Sobel filters to compute ed ge maps. These task-agnostic outputs are fused into multi-channel augmented images serving as the exclusive input for the Executor. The Executor, implemented as a DMPO agent with generic CNN and MLP encoders, predicts manipulation actions relying solely on these structural outlines without perceiving raw RGB pixels.}
  \label{fig:DE2}
\end{figure}

In this series of experiments, we introduce a generic, open-vocabulary Dispatcher based on the pre-trained OWL-ViT Vision Transformer \citep{10.1007/978-3-031-20080-9_42}. This approach moves beyond task-specific, handcrafted detectors, allowing the generic Dispatcher to semantically interpret the scene based on natural language queries. The Dispatcher receives RGB inputs from available scene overview cameras and processes them through two parallel streams: a structural stream utilizing standard Sobel filters to compute generic, low-level edge maps, and a semantic stream where OWL-ViT, prompted by one or more natural language keywords, generates high-level segmentation masks for specified objects.
These streams are fused into a strongly regularized communication channel represented as multi-channel augmented images for each view. The primary channel contains the raw, task-agnostic edge map, while subsequent channels contain this same edge map masked by the semantic segmentations corresponding to each generic text query. Consequently, the Executor—implemented as a DMPO agent with standard shallow generic encoders—never perceives raw RGB pixels; it relies solely on structural outlines highlighted by the Dispatcher according to the immediate semantic goal.

We validate this instantiation on a multi-object robotic manipulation task using the Aloha bi-manual platform. Four diverse fruits—a banana, pear, orange, and apple—are instantiated on the tabletop with randomized positions and orientations. The generic Dispatcher operates using only two fixed overview cameras (overhead and "worm's-eye" generic views), without recourse to wrist-mounted sensors. The generic agent must identify, reach, and lift a specific target fruit from the cluttered table, with the target specified solely via a natural language keyword passed to the Dispatcher. The task utilizes a staged generic reward function that incentivizes reaching the target prop and subsequently lifting it above a predefined height threshold.

\subsection{Experimental Evaluation: Semantic Decoupling and Geometric Transfer}

To validate the hypothesis that the D/E architecture effectively compartmentalizes semantic knowledge from mechanical control, we conducted a series of zero-shot transfer experiments. We compare our Open-Vocabulary D/E agent against a strong baseline: a standard, monolithic Multi-Task RL agent (end-to-end RGB) trained with the same MPO algorithm and network capacity.

Our experiments utilize four target objects with distinct shapes and textures: an Apple (round, red), an Orange (round, orange), a Banana (elongated, yellow), and a Pear (irregular/conical, green). We focus on three specific transfer scenarios:

\begin{table}[ht]
\centering
\caption{
Quantitative evaluation of zero-shot generalization.
We compare the success rates of the Standard RGB baseline and the proposed D/E agent across three types of distribution shifts: visual background changes (Rows 1-2), visual clutter (Rows 3-4), and semantic transfer to novel target objects (Rows 5-12).
}
\label{tab:owl_results}
\vspace{0.2cm}
\resizebox{\textwidth}{!}{%
\begin{tabular}{l l l l l c}
\hline
\textbf{Exp} & \textbf{Train Lift Task} & \textbf{Test Lift Task} & \textbf{Condition} & \textbf{Method} & \textbf{Success Rate} \\ \hline

1 & Apple & Apple diff. background & Zero-shot to different background & Standard RGB & 3.27\% \\
2 & Apple & Apple diff. background & Zero-shot to different background & D/E & 100\% \\
\hline
3 & Apple & Apple + clutter & Zero-shot with added clutter & Standard RGB & 7.62\% \\
4 & Apple & Apple + clutter & Zero-shot with added clutter & D/E & 98.71\% \\
\hline
5 & Apple & Orange & Zero-shot to similar shape diff. color & Standard RGB & 0\% \\
6 & Apple & Orange & Zero-shot to similar shape diff. color & D/E & 100\% \\
7 & Apple & Pear & Zero-shot to different shape & D/E & 91.67\% \\
8 & Apple & Banana & Zero-shot to very different shape & D/E & 16.24\% \\
\hline
9 & Banana & Pear & Zero-Shot to different shape & D/E & 87.94\% \\
10 & Apple $\lor$ Banana & Pear & Zero-Shot to different shape & D/E & 98.23\% \\
11 & Pear & Banana & Zero-shot to very different shape & D/E & 23.35\% \\
12 & Pear $\lor$ Apple & Banana & Zero-shot to very different shape & D/E & 40.02\% \\
\hline
\end{tabular}%
}
\end{table}

\textbf{1. Invariance to Visual Background Shift:}
We evaluated the ability to ignore irrelevant visual noise.
We train both the D/E agent and the Baseline agents on the ``Apple  lift'' task.
Both agents are trained in an environment having an ``Office'' background (\href{https://youtu.be/8ZskIr7zurA}{youtu.be/8ZskIr7zurA}). We then evaluated them zero-shot against a uniform dark blue background, keeping the robot and target object identical (\href{https://youtu.be/mzco\_zMMVSc}{youtu.be/mzco\texttt{\_}zMMVSc}).
The results (Table \ref{tab:owl_results}, Rows 1-2) demonstrate the standard RGB agent's inability to generalize under distribution shifts induced by background changes. The agent overfits to spurious correlations, illustrating a common failure mode of standard end-to-end learning.
The Baseline agent suffers a performance collapse as the success rate drops to 3.27\%.
In contrast, the D/E principle mandates a regularized communication channel that allows the performance to be unaltered by background noise. In the presented instantiation, the OWL-ViT based Dispatcher explicitly filters out task-irrelevant background information. This structural constraint ensures the Executor perceives the uniform background as functionally identical to the complex office scene, preserving high performance without the need for domain randomization or retraining.

\textbf{2. Robustness to Semantic Distractors:}
We evaluate the ability to ignore added clutter.
We take both agents trained on the ``Apple lift'' task and apply them without any additional training on a test environment where four fruits are placed in the workspace simultaneously (\href{[https://youtu.be/ClcoUzsIR0M](https://youtu.be/ClcoUzsIR0M)}{youtu.be/ClcoUzsIR0M}).
The D/E agent succeeds in the vast majority of cases (Table \ref{tab:owl_results}, Row 4). The Dispatcher handles the visual complexity by \textit{silencing} distractor objects. This ensures the Executor perceives a state functionally similar to a single-object scene, even if the target is partly occluded.
Consequently, performance remains high, whereas the success rate of the standard end-to-end baseline deteriorates significantly (Table \ref{tab:owl_results}, Row 3).

\textbf{3. Zero-Shot Generalization to Novel Objects:}
We evaluate the ability to transfer learned manipulation behaviors to semantically similar contexts. Specifically, target objects with similar geometry but different visual appearance (e.g., color and shape). 
We apply both agents, originally trained on the ``Apple lift'' task, to an ``Orange lift'' task without any additional training (\href{https://youtu.be/OlH8PA9Q6IU}{youtu.be/OlH8PA9Q6IU}).
For the D/E agent, adaptation requires only updating the text prompt sent to the Dispatcher from ``Apple'' to ``Orange''.
In contrast, the Baseline agent, having overfitted to the specific texture and color of the apple, fails completely, achieving a 0\% success rate on the orange (Table \ref{tab:owl_results}, Row 5).
The D/E Executor, however, receives regularized input; since the semantic mask of the orange is geometrically similar to that of the apple, the Dispatcher treats it as a known object. This results in a 100\% success rate without fine-tuning (Table \ref{tab:owl_results}, Row 6).

We extend this evaluation by performing additional zero-shot tests on target objects with increasing geometric dissimilarity from the target object used during training (Table \ref{tab:owl_results}, Rows 7-8).
The D/E method, achieves a 91.67\% zero-shot success rate on the ``Pear lift'' task, despite the pear's distinct color and only partial geometric similarity to the apple (\href{https://youtu.be/cu__ZcXYi4k}{youtu.be/cu\texttt{\_}\texttt{\_}ZcXYi4k}).
The ``Banana lift'' task achieves a 16.24\% success rate (\href{https://youtu.be/k2k2lqif_TE}{youtu.be/k2k2lqif\texttt{\_}TE}), even though the banana shares neither texture nor geometric similarity with the original target object.
As expected, zero-shot performance correlates with the degree of geometric similarity to the original training object.
Crucially, the D/E approach enables the retention of some success capabilities even when facing significantly different target objects.

\begin{figure}[t]
    \centering
    \begin{minipage}[t]{0.49\textwidth}
        \centering
        \includegraphics[width=\textwidth]{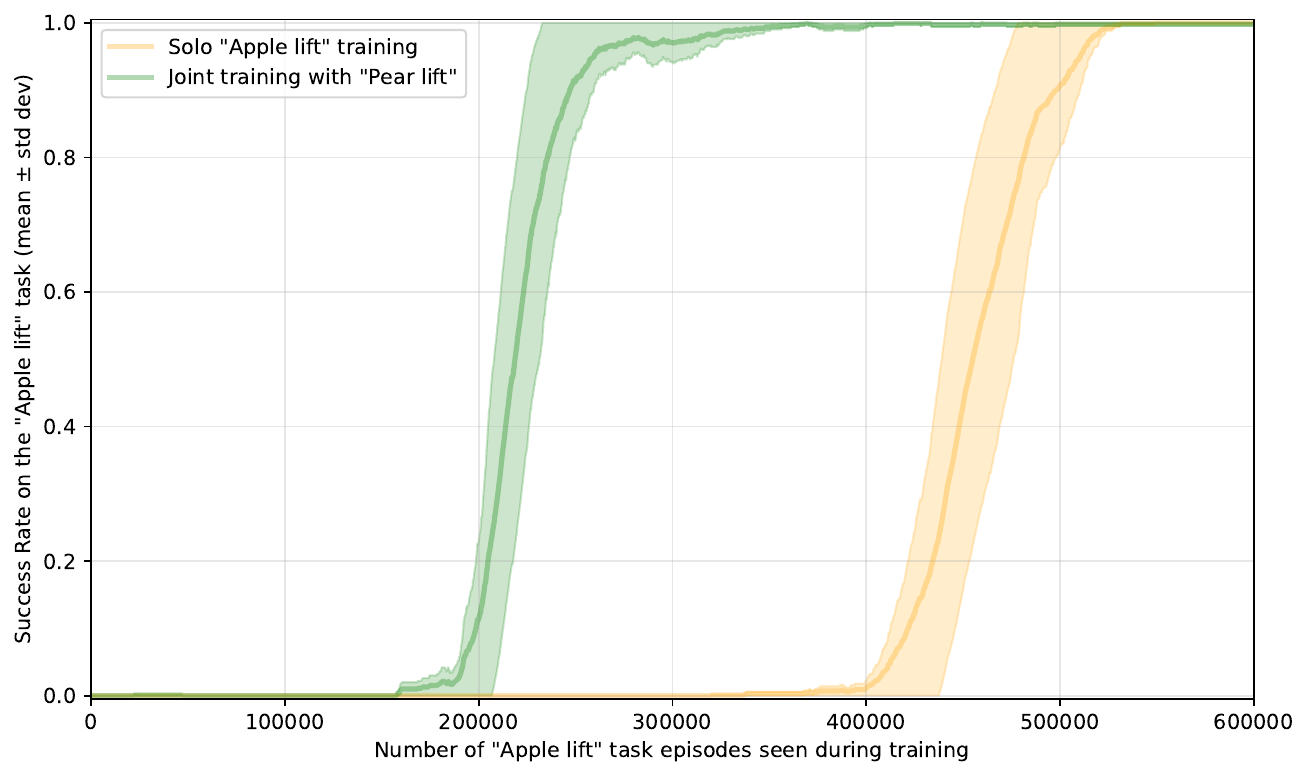}
        \caption{Training efficiency comparison on the ``Apple lift'' task. The plot reports the success rate (mean $\pm$ standard deviation) as a function of the number of training episodes. The agent trained jointly with the ``Pear lift'' task (green) converges significantly faster than the agent trained on ``Apple lift'' in isolation (orange), validating the data efficiency achieved by the shared Executor.}
        \label{fig:plot_left}
    \end{minipage}
    \hfill 
    \begin{minipage}[t]{0.49\textwidth}
        \centering
        \includegraphics[width=\textwidth]{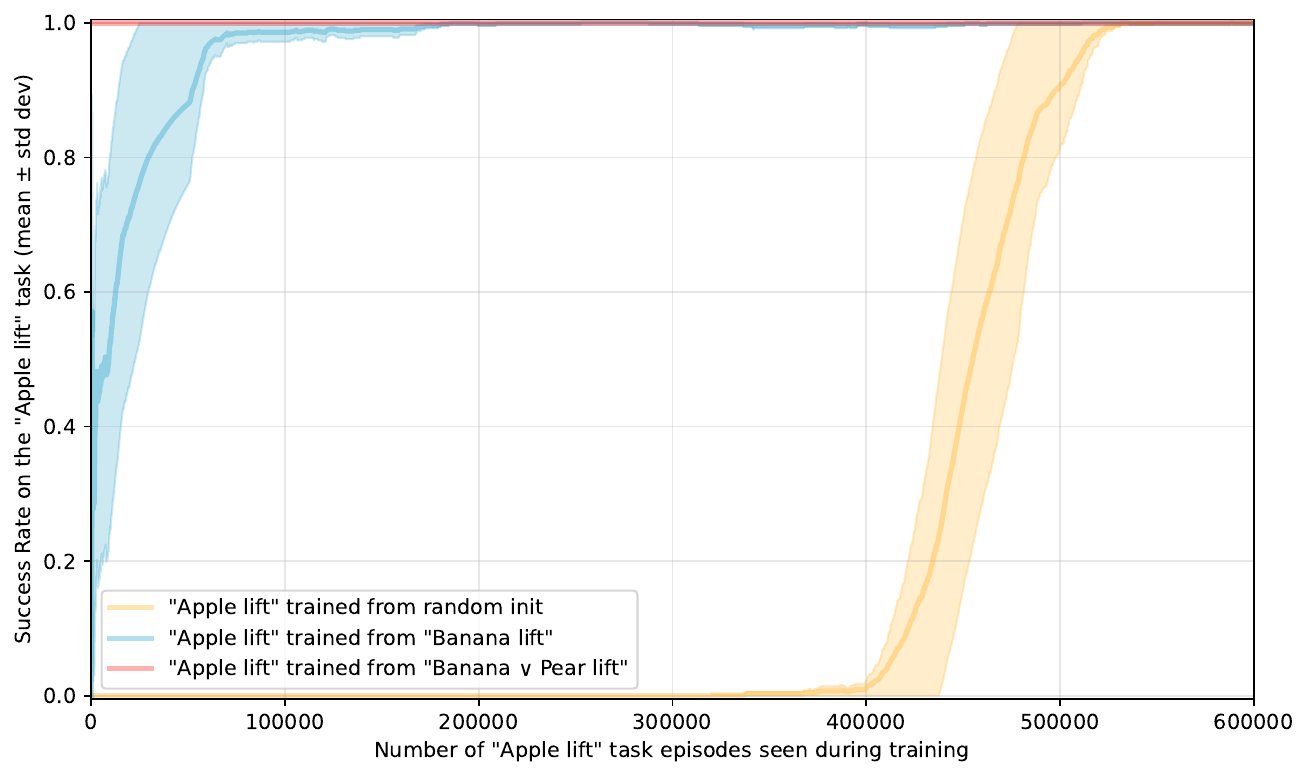}
        \caption{Sequential transfer efficiency on the ``Apple lift'' task. We compare training from random initialization (orange) against fine-tuning from a policy pre-trained on ``Banana lift'' (blue) and a policy pre-trained on both ``Banana lift'' and ``Pear lift'' (red). Pre-training on a single task reduces sample complexity by over 50\%, while pre-training on two tasks yields immediate 100\% zero-shot success.}
        \label{fig:plot_right}
    \end{minipage}
\end{figure}

\textbf{4. Generalization and Efficiency via Multi-Task Learning}

We aim to demonstrate empirically that the Dispatcher and Executor modules acquire generic, reusable knowledge. Specifically, we hypothesize that an Executor shared across a higher number of tasks performed by the same embodiment acquires a more general representation of procedural (``how'') knowledge. We measure this generality through two metrics: training efficiency and zero-shot performance on unseen scenarios.

\textbf{Training Efficiency.}
First, we train a shared Executor (from random initialization) jointly on two distinct tasks: ``Apple lift'' and ``Banana lift''. While the Executor is shared, each task utilizes a specific OWL-ViT Dispatcher conditioned on a distinct text query (``\textit{Apple}'' or ``\textit{Banana}''). We limit the joint training to the same total computational budget used for the single-task ``Apple lift'' baseline.
Consequently, although each task in the joint setup effectively sees only half the training samples, both achieve a 100\% success rate.
In contrast, training the ``Apple lift'' policy in isolation with this reduced sample size fails to converge (Figure \ref{fig:plot_left}).
This confirms that a shared Executor improves data efficiency by acting as a repository for embodiment-specific knowledge that is transferred across tasks.

\textbf{Zero-Shot Generalization.}
We further test the hypothesis that sharing the Executor improves zero-shot generalization to unseen objects. We compare three policies trained from random initialization with identical compute budgets:
1) Trained solely on ``Apple lift'';
2) Trained solely on ``Banana lift'';
3) Trained jointly on ``Apple lift'' and Banana ``lift'', forcing the Executor to learn grasping primitives for both round and elongated geometries.
We evaluate these policies zero-shot on the ``Pear lift'' task. As expected, the multi-task Executor achieves superior performance (Table \ref{tab:owl_results}, Rows 7, 9, 10). Transferring from the single-task ``Apple'' and ``Banana'' policies yields success rates of 91.67\% and 87.94\%, respectively. In contrast, the jointly trained Executor achieves 98.23\%, reducing the error rate by more than a factor of four.
To validate the generality of this phenomenon, we repeat the experiment targeting the banana for zero-shot evaluation, using policies trained on apples and pears (Table \ref{tab:owl_results}, Rows 8, 11, 12). While single-task transfer achieves only 16.24\% and 23.35\%, the jointly trained Executor reaches 40.02\%, providing further empirical support for the proposed hypothesis.

\textbf{Sequential Transfer Efficiency.} Finally, we investigate the benefits of the shared Executor in a sequential transfer setting, distinguishing between learning from scratch and fine-tuning from pre-existing knowledge.
We compare three training regimes for the ``Apple lift'' task: 1) training from random initialization, 2) fine-tuning from a policy pre-trained on the ``Banana lift'' task, and 3) fine-tuning from a policy pre-trained on both ``Banana lift'' and ``Pear lift'' tasks jointly.
The results are illustrated in Figure \ref{fig:plot_right}.
The baseline policy (random initialization) requires approximately 550,000 episodes to reach convergence.
In contrast, fine-tuning from a single related task (``Banana lift'') dramatically accelerates learning, converging in less than half the number of samples. Most notably, the policy pre-trained on two tasks (``Banana'' and ``Pear'') demonstrates perfect immediate generalization, achieving a 100\% success rate from the start of training.
This progression—from slow learning, to rapid adaptation, to immediate zero-shot execution—provides compelling evidence that the Executor accumulates robust, reusable procedural knowledge as it is exposed to a wider variety of tasks.




Collectively, these experiments validate the core premise of the Dispatcher/Executor principle: that effective multi-task control relies on the clean separation of semantic intent from mechanical execution. By decoupling the ``what'' (Dispatcher) from the ``how'' (Executor), we break the dependency between perceiving a new context and learning the actuation primitives to act within it.
As demonstrated, this architecture not only ensures robustness against visual distractors and background shifts but also allows the agent to apply existing control primitives to novel scenarios immediately upon identification. 
Consequently, the D/E framework effectively reduces the marginal sample complexity of learning new tasks, offering a scalable path toward general-purpose robotic behavior learning.

\section{Discussion}

This position paper introduces the Dispatcher/Executor (D/E) principle for structuring control architectures in multi-task reinforcement learning. It advocates for: a) separating the architecture into a Dispatcher module, which holds general world knowledge and understands the task; b) an Executor module, which learns specific interactions with a particular device; and c) a regularizing communication channel that enforces abstract and compositional messaging between them.

We presented a concrete implementation of such a D/E architecture for robotic manipulation. In two distinct learning scenarios—reinforcement learning from experience and hindsight transfer—we demonstrated a substantial improvement in generalization. Using the same budget of device interactions, the D/E architecture successfully transferred its control abilities to an entire class of related tasks.

While our current implementation relies on certain engineered features, future work will focus on end-to-end learning architectures based on the D/E principle. The separation of control based on the type of knowledge required positions the Dispatcher as an ideal candidate for integrating Large Multi-modal Models (LMMs). Their capacity for general world reasoning and tool use aligns perfectly with the need to translate high-level task descriptions into tailored Executor calls. Additionally, a key subject for future research is learning regularized representations that discover optimal abstractions for the communication channel between Dispatcher and Executor.

\section*{Acknowledgments}
Thanks to the Control Team and various colleagues at DeepMind for ongoing discussions.


\bibliography{main}  
\bibliographystyle{rlc}

\end{document}